\documentclass[authoryear, hidelinks]{elsarticle}\usepackage[]{graphicx}\usepackage[]{color}
\makeatletter
\def\maxwidth{ %
  \ifdim\Gin@nat@width>\linewidth
    \linewidth
  \else
    \Gin@nat@width
  \fi
}
\makeatother

\definecolor{fgcolor}{rgb}{0.345, 0.345, 0.345}

\usepackage{framed}
\makeatletter
 {\par\unskip\endMakeFramed%
 \at@end@of@kframe}
\makeatother

\definecolor{shadecolor}{rgb}{.97, .97, .97}
\definecolor{messagecolor}{rgb}{0, 0, 0}
\definecolor{warningcolor}{rgb}{1, 0, 1}
\definecolor{errorcolor}{rgb}{1, 0, 0}

\usepackage{alltt}
\makeatletter
\def\ps@pprintTitle{%
 \let\@oddhead\@empty
 \let\@evenhead\@empty
 \def\@oddfoot{}%
 \let\@evenfoot\@oddfoot}
\makeatother

\usepackage{natbib}
\pdfoutput=1
\usepackage{hyperref}
\hypersetup{
    colorlinks=false,
    linkcolor=black,
    citecolor=black,
    filecolor=black,
    urlcolor=black,
}
\definecolor{shadecolor}{rgb}{.97, .97, .97}
\definecolor{messagecolor}{rgb}{0, 0, 0}
\definecolor{warningcolor}{rgb}{1, 0, 1}
\definecolor{errorcolor}{rgb}{1, 0, 0}

\usepackage{alltt}
\usepackage[utf8]{inputenc}
\usepackage[english]{babel}
\usepackage[a4paper,
            left=1.4in,right=1.4in,top=1.4in,bottom=1.4in
            ]{geometry}

\usepackage{amsthm,amsmath,amssymb,amstext}
\usepackage{booktabs}
\usepackage{array}
\usepackage{url}

\makeatletter
\def\maxwidth{ %
  \ifdim\Gin@nat@width>\linewidth
    \linewidth
  \else
    \Gin@nat@width
  \fi
}
\makeatother

\usepackage{etoolbox}
\appto\UrlBreaks{\do\a\do\b\do\c\do\d\do\e\do\f\do\g\do\h\do\i\do\j
\do\k\do\l\do\m\do\n\do\o\do\p\do\q\do\r\do\s\do\t\do\u\do\v\do\w
\do\x\do\y\do\z}

\usepackage{nicefrac}
\usepackage{caption}
\usepackage{tabularx}
\usepackage{graphicx}
\usepackage{color}
\usepackage{booktabs}
\usepackage{lscape}
\usepackage{placeins}
\usepackage{colortbl}
\usepackage{dsfont}
\usepackage{multirow}
\usepackage{listings}
\usepackage{adjustbox}
\usepackage{setspace}

\usepackage[colorinlistoftodos,prependcaption,textsize=tiny]{todonotes}
\IfFileExists{upquote.sty}{\usepackage{upquote}}{}
\IfFileExists{upquote.sty}{\usepackage{upquote}}{}
\begin{document}

\title{Automatic Exploration of Machine Learning Experiments on OpenML}

\author[1]{Daniel Kühn*}
\ead{daniel.kuehn.87@gmail.com}
\author[1]{Philipp Probst*}
\ead{philipp\_probst@gmx.de}
\author[1]{Janek Thomas}
\ead{janek.thomas@stat.uni-muenchen.de}
\author[1]{Bernd Bischl}
\ead{bernd\_bischl@gmx.net}

\address[1]{Ludwig-Maximilians-Universit\"at M\"unchen, Germany}

\begin{abstract}
Understanding the influence of hyperparameters on the performance of a machine learning algorithm is an important scientific topic in itself and can help to improve automatic hyperparameter tuning procedures.
Unfortunately, experimental meta data for this purpose is still rare.
This paper presents a large, free and open dataset addressing this problem, containing results
on 38 OpenML data sets, six different machine learning algorithms and many different hyperparameter configurations.
Results where generated by an automated random sampling strategy, termed the \textit{OpenML Random Bot}.
Each algorithm was cross-validated up to $20.000$ times per dataset with different hyperparameters settings, resulting in a meta dataset of around $2.5$ million experiments overall.


\end{abstract}

\maketitle

\section{Introduction}

When applying machine learning algorithms on real world datasets, users have to choose from a large selection of different algorithms with many of them offering a set of hyperparameters to control algorithmic performance.
Although sometimes default values exist, there is no agreed upon principle for their definition
(but see our recent work in in \citep{Probst2018} for a potential approach).
Automatic tuning of such parameters is a possible solution \citep{Claesen2015}, but comes with a considerable computational burden.



Meta-learning tries to decrease this cost \citep{Feurer2015}, by reusing information of previous runs of the algorithm on similar datasets, which obviously requires access to such prior empirical results.
With this paper we provide a freely accessible meta dataset that contains around $2.5$ million runs of six different machine learning algorithms on $38$ classification datasets.

Large, freely available datasets like Imagenet \citep{ImageNet2009} are important for the progress of machine learning, we hope to support developments in the area of meta-learning and benchmarking, meta-learning and hyperparameter tuning with our work here.

While similar meta-datasets have been created in the past, we were not able to access them by the links provided in their respective papers:
\citet{Smith2014} provides a repository with Weka-based machine learning experiments on 72 data sets, 9 machine learning algorithms, 10 hyperparameter settings for each algorithm, and several meta-features of each data set.
\citet{Reif2012ACD} created a meta-dataset based on machine learning experiments on 83 datasets, 6 classification algorithms, and 49 meta features.

In this paper, we describe our experimental setup, specify how our meta-dataset is created by running random machine learning experiments through the OpenML platform \citep{OpenML2013} and explain how to access our results.

\section{Considered ML data sets, algorithms and hyperparameters}

To create the meta dataset, six supervised machine learning algorithms are run on 38 classification tasks.
For each algorithm the available hyperparameters are explored in a predefined range (see Table~\ref{tab:parameter}).
Some of these hyperparameters are transformed by the function found in column \textit{trafo} of Table~\ref{tab:parameter}
to allow non-uniform sampling, a usual procedure in tuning.


\begin{table}[ht]
\centering
\begin{tabular}{llrrrr}
 algorithm & hyperparameter & type & lower & upper & trafo \\ 
  \hline
glmnet & alpha & numeric & 0 & 1 & - \\ 
   & lambda & numeric & -10 & 10 & $2^x$ \\ 
   \hline
rpart & cp & numeric & 0 & 1 & - \\ 
   & maxdepth & integer & 1 & 30 & - \\ 
   & minbucket & integer & 1 & 60 & - \\ 
   & minsplit & integer & 1 & 60 & - \\ 
   \hline
kknn & k & integer & 1 & 30 & - \\ 
   \hline
svm & kernel & discrete & - & - & - \\ 
   & cost & numeric & -10 & 10 & $2^x$ \\ 
   & gamma & numeric & -10 & 10 & $2^x$ \\ 
   & degree & integer & 2 & 5 & - \\ 
   \hline
ranger & num.trees & integer & 1 & 2000 & - \\ 
   & replace & logical & - & - & - \\ 
   & sample.fraction & numeric & 0 & 1 & - \\ 
   & mtry & numeric & 0 & 1 & $x \cdot p$ \\ 
   & respect.unordered.factors & logical & - & - & - \\ 
   & min.node.size & numeric & 0 & 1 & $n^x$ \\ 
   \hline
xgboost & nrounds & integer & 1 & 5000 & - \\ 
   & eta & numeric & -10 & 0 & $2^x$ \\ 
   & subsample & numeric & 0 & 1 & - \\ 
   & booster & discrete & - & - & - \\ 
   & max\_depth & integer & 1 & 15 & - \\ 
   & min\_child\_weight & numeric & 0 & 7 & $2^x$ \\ 
   & colsample\_bytree & numeric & 0 & 1 & - \\ 
   & colsample\_bylevel & numeric & 0 & 1 & - \\ 
   & lambda & numeric & -10 & 10 & $2^x$ \\ 
   & alpha & numeric & -10 & 10 & $2^x$ \\ 
   \hline
\end{tabular}
\caption{Hyperparameters of the algorithms. $p$ refers to the number of variables and $n$ to the
    number of observations. The used algorithms are  \texttt{glmnet} \citep{glmnet}, \texttt{rpart} \citep{rpart}, \texttt{kknn} \citep{kknn},  \texttt{svm} \citep{svm}, \texttt{ranger} \citep{ranger} and \texttt{xgboost} \citep{xgboost}.} 
\label{tab:parameter}
\end{table}

These algorithms are run on a subset of the OpenML100 benchmark suite \citep{Bischl2017}, which consists of 100 classification datasets, carefully curated from the thousands of datasets available on OpenML \citep{OpenML2013}.
We only include datasets without missing data and with a binary outcome resulting in 38 datasets.
The datasets and their respective characteristics can be found in Table~\ref{tab:datasets}.

\begin{table}[ht]
\centering
\begingroup\scriptsize
\begin{tabular}{rrlrrrrr}
  \hline
Data\_id & Task\_id & Name & n & p & majPerc & numFeat & catFeat \\ 
  \hline
  3 &   3 & kr-vs-kp & 3196 &  37 & 0.52 &   0 &  37 \\ 
   31 &  31 & credit-g & 1000 &  21 & 0.70 &   7 &  14 \\ 
   37 &  37 & diabetes & 768 &   9 & 0.65 &   8 &   1 \\ 
   44 &  43 & spambase & 4601 &  58 & 0.61 &  57 &   1 \\ 
   50 &  49 & tic-tac-toe & 958 &  10 & 0.65 &   0 &  10 \\ 
  151 & 219 & electricity & 45312 &   9 & 0.58 &   7 &   2 \\ 
  312 & 3485 & scene & 2407 & 300 & 0.82 & 294 &   6 \\ 
  333 & 3492 & monks-problems-1 & 556 &   7 & 0.50 &   0 &   7 \\ 
  334 & 3493 & monks-problems-2 & 601 &   7 & 0.66 &   0 &   7 \\ 
  335 & 3494 & monks-problems-3 & 554 &   7 & 0.52 &   0 &   7 \\ 
  1036 & 3889 & sylva\_agnostic & 14395 & 217 & 0.94 & 216 &   1 \\ 
  1038 & 3891 & gina\_agnostic & 3468 & 971 & 0.51 & 970 &   1 \\ 
  1043 & 3896 & ada\_agnostic & 4562 &  49 & 0.75 &  48 &   1 \\ 
  1046 & 3899 & mozilla4 & 15545 &   6 & 0.67 &   5 &   1 \\ 
  1049 & 3902 & pc4 & 1458 &  38 & 0.88 &  37 &   1 \\ 
  1050 & 3903 & pc3 & 1563 &  38 & 0.90 &  37 &   1 \\ 
  1063 & 3913 & kc2 & 522 &  22 & 0.80 &  21 &   1 \\ 
  1067 & 3917 & kc1 & 2109 &  22 & 0.85 &  21 &   1 \\ 
  1068 & 3918 & pc1 & 1109 &  22 & 0.93 &  21 &   1 \\ 
  1120 & 3954 & MagicTelescope & 19020 &  12 & 0.65 &  11 &   1 \\ 
  1461 & 14965 & bank-marketing & 45211 &  17 & 0.88 &   7 &  10 \\ 
  1462 & 10093 & banknote-authentication & 1372 &   5 & 0.56 &   4 &   1 \\ 
  1464 & 10101 & blood-transfusion-service-center & 748 &   5 & 0.76 &   4 &   1 \\ 
  1467 & 9980 & climate-model-simulation-crashes & 540 &  21 & 0.91 &  20 &   1 \\ 
  1471 & 9983 & eeg-eye-state & 14980 &  15 & 0.55 &  14 &   1 \\ 
  1479 & 9970 & hill-valley & 1212 & 101 & 0.50 & 100 &   1 \\ 
  1480 & 9971 & ilpd & 583 &  11 & 0.71 &   9 &   2 \\ 
  1485 & 9976 & madelon & 2600 & 501 & 0.50 & 500 &   1 \\ 
  1486 & 9977 & nomao & 34465 & 119 & 0.71 &  89 &  30 \\ 
  1487 & 9978 & ozone-level-8hr & 2534 &  73 & 0.94 &  72 &   1 \\ 
  1489 & 9952 & phoneme & 5404 &   6 & 0.71 &   5 &   1 \\ 
  1494 & 9957 & qsar-biodeg & 1055 &  42 & 0.66 &  41 &   1 \\ 
  1504 & 9967 & steel-plates-fault & 1941 &  34 & 0.65 &  33 &   1 \\ 
  1510 & 9946 & wdbc & 569 &  31 & 0.63 &  30 &   1 \\ 
  1570 & 9914 & wilt & 4839 &   6 & 0.95 &   5 &   1 \\ 
  4134 & 14966 & Bioresponse & 3751 & 1777 & 0.54 & 1776 &   1 \\ 
  4534 & 34537 & PhishingWebsites & 11055 &  31 & 0.56 &   0 &  31 \\ 
   \hline
\end{tabular}
\endgroup
\caption{Included datasets and respective characteristics. \textit{n} are the number of observations, \textit{p} the number of features, \textit{maj.class} the percentage of observations in the largest class, \textit{numFeat} the number of numeric features and \textit{catFeat} the number of categorical features.} 
\label{tab:datasets}
\end{table}

\newpage

\section{Random Experimentation Bot}

To conduct a large number of experiments a bot was implemented to automatically plan and execute runs, following the paradigm of random search.
The bot iteratively executes these steps: 

\begin{enumerate}
\item Randomly sample a task $T$ (with an associated data set) from Table~\ref{tab:datasets}.
\item Randomly sample one ML algorithm $A$.
\item Randomly sample a hyperparameter setting $\theta$ of algorithm $A$, uniformly from the ranges specified in Table~\ref{tab:parameter},
then transform, if a transformation function is given.
\item Obtain task $T$ (and dataset) from OpenML and store it locally.
\item Evaluate algorithm $A$ with configuration $\theta$ on task $T$, with associated 10-fold cross-validation from OpenML.
\item Upload run results to OpenML, including hyperparameter configuration and time measurements.
\item OpenML now calculates various performance metrics for the uploaded cross-validated predictions.
\item The OpenML-ID of the bot (2702)  and the tag \texttt{mlrRandomBot} is used for identification.
\end{enumerate}

A clear advantage of random sampling is that all bot runs are completely independent of each other, making all experiments embarrassingly parallel.
Furthermore, more experiments can easily and conveniently added later on, without introducing any kind of bias into the sampling method.

The bot is developed open source in R and can be found on GitHub\footnote{\url{https://github.com/ja-thomas/OMLbots}}.
The bot is based on the R packages \texttt{mlr} \citep{Bischl2016} and \texttt{OpenML} \citep{Casalicchio2017} and written in modular form such that it can be extended with new sampling strategies for hyperparameters, algorithms and datasets in the future. Parallelization was performed with R package \texttt{batchtools} \citep{Lang2017}.

After more than $6$ million benchmark experiments the results of the bot are downloaded from OpenML.
For each of the algorithms $500000$ experiments are used to obtain the final dataset.
The experiments are chosen by the following procedure: For each algorithm, a threshold $B$ is set (see below) and, if the number of results for a dataset exceeds $B$, we draw randomly $B$ of the results obtained for this algorithm and this dataset. The threshold value $B$ is chosen for each algorithm separately to exactly obtain in total 500000 results for each algorithm.

For \texttt{kknn} we only execute 30 experiments per dataset because this number of experiments is high enough
to cover the hyperparameter space (that only consists of the parameter $k$ for $k \in \{1,...,30\}$) appropriately, resulting in 1140 experiments.
All in all this results in around 2.5 million experiments.

The distribution of the runs on the datasets and algorithms is displayed in Table~\ref{tab:datasets2}.

\begin{table}[ht]
\centering
\begingroup\scriptsize
\begin{tabular}{lrrrrrrrr}
  \hline
Data\_id & Task\_id & glmnet & rpart & kknn & svm & ranger & xgboost & Total \\ 
  \hline
3 & 3 & 15547 & 14633 & 30 & 19644 & 15139 & 16867 & 81860 \\ 
  31 & 31 & 15547 & 14633 & 30 & 19644 & 15139 & 16867 & 81860 \\ 
  37 & 37 & 15546 & 14633 & 30 & 15985 & 15139 & 16866 & 78199 \\ 
  44 & 43 & 15547 & 14633 & 30 & 19644 & 15139 & 16867 & 81860 \\ 
  50 & 49 & 15547 & 14633 & 30 & 19644 & 15139 & 16866 & 81859 \\ 
  151 & 219 & 15547 & 14632 & 30 & 2384 & 12517 & 16866 & 61976 \\ 
  312 & 3485 & 6613 & 13455 & 30 & 18740 & 12985 & 15886 & 67709 \\ 
  333 & 3492 & 15546 & 14632 & 30 & 19644 & 15139 & 16867 & 81858 \\ 
  334 & 3493 & 15547 & 14633 & 30 & 19644 & 14492 & 16867 & 81213 \\ 
  335 & 3494 & 15547 & 14633 & 30 & 15123 & 15139 & 10002 & 70474 \\ 
  1036 & 3889 & 14937 & 14633 & 30 & 2338 & 7397 & 2581 & 41916 \\ 
  1038 & 3891 & 15547 & 5151 & 30 & 5716 & 4827 & 1370 & 32641 \\ 
  1043 & 3896 & 6466 & 14633 & 30 & 10121 & 3788 & 16867 & 51905 \\ 
  1046 & 3899 & 15547 & 14633 & 30 & 5422 & 8842 & 11812 & 56286 \\ 
  1049 & 3902 & 7423 & 14632 & 30 & 12064 & 15139 & 4453 & 53741 \\ 
  1050 & 3903 & 15547 & 14633 & 30 & 19644 & 11357 & 13758 & 74969 \\ 
  1063 & 3913 & 15547 & 14633 & 30 & 19644 & 7914 & 16866 & 74634 \\ 
  1067 & 3917 & 15546 & 14632 & 30 & 10229 & 7386 & 16866 & 64689 \\ 
  1068 & 3918 & 15546 & 14633 & 30 & 13893 & 8173 & 16866 & 69141 \\ 
  1120 & 3954 & 15531 & 7477 & 30 & 3908 & 9760 & 8143 & 44849 \\ 
  1461 & 14965 & 6970 & 14073 & 30 & 2678 & 14323 & 2215 & 40289 \\ 
  1462 & 10093 & 8955 & 14633 & 30 & 6320 & 15139 & 16867 & 61944 \\ 
  1464 & 10101 & 15547 & 14632 & 30 & 19644 & 15139 & 16867 & 81859 \\ 
  1467 & 9980 & 15547 & 14633 & 30 & 4441 & 15139 & 16866 & 66656 \\ 
  1471 & 9983 & 15547 & 14633 & 30 & 9725 & 13523 & 16866 & 70324 \\ 
  1479 & 9970 & 15546 & 14633 & 30 & 19644 & 15140 & 16867 & 81860 \\ 
  1480 & 9971 & 15024 & 14633 & 30 & 19644 & 15139 & 16254 & 80724 \\ 
  1485 & 9976 & 8247 & 10923 & 30 & 10334 & 15139 & 9237 & 53910 \\ 
  1486 & 9977 & 3866 & 11389 & 30 & 1490 & 15139 & 5813 & 37727 \\ 
  1487 & 9978 & 15547 & 6005 & 30 & 19644 & 15139 & 11194 & 67559 \\ 
  1489 & 9952 & 15547 & 14633 & 30 & 17298 & 15139 & 16867 & 79514 \\ 
  1494 & 9957 & 15547 & 14632 & 30 & 19644 & 15139 & 16867 & 81859 \\ 
  1504 & 9967 & 15547 & 14633 & 30 & 19644 & 15140 & 16867 & 81861 \\ 
  1510 & 9946 & 15547 & 14633 & 30 & 19644 & 15139 & 16867 & 81860 \\ 
  1570 & 9914 & 15546 & 14632 & 30 & 19644 & 15139 & 16867 & 81858 \\ 
  4134 & 14966 & 1493 & 3947 & 30 & 560 & 14516 & 2222 & 22768 \\ 
  4534 & 34537 & 2801 & 3231 & 30 & 2476 & 15139 & 947 & 24624 \\ 
   \hline
Total & 257661 & 486995 & 485368 & 1110 & 485549 & 484860 & 486953 & 2430835 \\ 
   \hline
\end{tabular}
\endgroup
\caption{Number of experiments for each combination of dataset and algorithm.} 
\label{tab:datasets2}
\end{table}

\newpage

\section{Access to the results}
The results of the benchmark can be accessed in different ways:

\begin{itemize}
\item The easiest way to access them is to go to the figshare repository \citep{Kuehn2018} and to download the \texttt{.csv} files. For each algorithm there is a csv file that contains a row for each algorithm run with the columns \texttt{Data\_id}, the hyperparameter settings, the performance measures (auc, accuracy and brier score), the runtime, the scimark reference runtime and some characteristics of the dataset such as  the number of features or the number of observations.
\item Alternatively the code for the extraction of the data from the nightly database snapshot of OpenML can
be found here: \url{https://github.com/ja-thomas/OMLbots/blob/master/snapshot_database/database_extraction.R}. With this script all results that were created by the random bot (OpenML-ID 2702) are downloaded and the final dataset is created. (Warning: As the OpenML database is updated daily, changes can occur.)
\end{itemize}

\section{Discussion and potential usage of the results}

The presented data can be used to study the effect and influence of hyperparameter setting on performance in various ways.
Possible applications are:
\begin{itemize}
\item Obtaining defaults for ML algorithm that work well across many datasets \citep{Probst2018};
\item Measuring the importance of hyperparameters, to investigate which should be tuned \citep[see][]{Rijn2017, Probst2018};
\item Obtaining ranges or priors of tuning parameters to focus on important regions of the search space \citep[see][]{Rijn2017, Probst2018};
\item Meta-Learning;
\item Investigating, debugging and improving the robustness of algorithms.
\end{itemize}

Possible weaknesses of the approach, which we would like to address in the future, are:
\begin{itemize}
\item For each ML algorithm, a set of considered hyperparameters and their initial ranges has to be provided. It would be much more convenient if the bot could handle the set of all technical hyperparameters, with infinite ranges.
\item Smarter, sequential sampling might be required to scale to high-dimensional hyperparameter spaces.
But note that we not only care about optimal configurations but much rather would like to learn as much as possible about the considered parameter space, including areas of bad performance.
So simply switching to Bayesian optimization or related search techniques might not be appropriate.
\end{itemize}



\bibliography{bot}

\begin{thebibliography}{19}
\providecommand{\natexlab}[1]{#1}
\providecommand{\url}[1]{\texttt{#1}}
\expandafter\ifx\csname urlstyle\endcsname\relax
  \providecommand{\doi}[1]{doi: #1}\else
  \providecommand{\doi}{doi: \begingroup \urlstyle{rm}\Url}\fi

\bibitem[Bischl et~al.(2016)Bischl, Lang, Kotthoff, Schiffner, Richter,
  Studerus, Casalicchio, and Jones]{Bischl2016}
B.~Bischl, M.~Lang, L.~Kotthoff, J.~Schiffner, J.~Richter, E.~Studerus,
  G.~Casalicchio, and Z.~M. Jones.
\newblock mlr: Machine learning in {R}.
\newblock \emph{Journal of Machine Learning Research}, 17\penalty0
  (170):\penalty0 1--5, 2016.

\bibitem[{Bischl} et~al.(2017){Bischl}, {Casalicchio}, {Feurer}, {Hutter},
  {Lang}, {Mantovani}, {van Rijn}, and {Vanschoren}]{Bischl2017}
B.~{Bischl}, G.~{Casalicchio}, M.~{Feurer}, F.~{Hutter}, M.~{Lang}, R.~G.
  {Mantovani}, J.~N. {van Rijn}, and J.~{Vanschoren}.
\newblock {{OpenML} benchmarking suites and the {OpenML100}}.
\newblock \emph{ArXiv preprint arXiv:1708.03731}, Aug. 2017.
\newblock URL \url{https://arxiv.org/abs/1708.03731}.

\bibitem[Casalicchio et~al.(2017)Casalicchio, Bossek, Lang, Kirchhoff,
  Kerschke, Hofner, Seibold, Vanschoren, and Bischl]{Casalicchio2017}
G.~Casalicchio, J.~Bossek, M.~Lang, D.~Kirchhoff, P.~Kerschke, B.~Hofner,
  H.~Seibold, J.~Vanschoren, and B.~Bischl.
\newblock {{OpenML}: An {R} package to connect to the machine learning platform
  {OpenML}}.
\newblock \emph{Computational Statistics}, 32\penalty0 (3):\penalty0 1--15,
  2017.

\bibitem[Chen and Guestrin(2016)]{xgboost}
T.~Chen and C.~Guestrin.
\newblock {XGBoost}: A scalable tree boosting system.
\newblock In \emph{Proceedings of the 22nd ACM SIGKDD International Conference
  on Knowledge Discovery and Data Mining}, KDD '16, pages 785--794, New York,
  NY, USA, 2016. ACM.

\bibitem[Claesen and Moor(2015)]{Claesen2015}
M.~Claesen and B.~D. Moor.
\newblock Hyperparameter search in machine learning.
\newblock \emph{MIC 2015: The XI Metaheuristics International Conference},
  2015.

\bibitem[Deng et~al.(2009)Deng, Dong, Socher, jia Li, Li, and
  Fei-fei]{ImageNet2009}
J.~Deng, W.~Dong, R.~Socher, L.~jia Li, K.~Li, and L.~Fei-fei.
\newblock Imagenet: A large-scale hierarchical image database.
\newblock \emph{IEEE Conference on Computer Vision and Pattern Recognition},
  pages 248--255, 2009.

\bibitem[Feurer et~al.(2015)Feurer, Springenberg, and Hutter]{Feurer2015}
M.~Feurer, J.~T. Springenberg, and F.~Hutter.
\newblock Initializing bayesian hyperparameter optimization via meta-learning.
\newblock In \emph{Proceedings of the Twenty-Ninth AAAI Conference on
  Artificial Intelligence}, pages 1128--1135. AAAI Press, 2015.

\bibitem[Friedman et~al.(2010)Friedman, Hastie, and Tibshirani]{glmnet}
J.~Friedman, T.~Hastie, and R.~Tibshirani.
\newblock Regularization paths for generalized linear models via coordinate
  descent.
\newblock \emph{Journal of Statistical Software}, 33\penalty0 (1):\penalty0
  1--22, 2010.

\bibitem[Kühn et~al.(2018)Kühn, Probst, Thomas, and Bischl]{Kuehn2018}
D.~Kühn, P.~Probst, J.~Thomas, and B.~Bischl.
\newblock {{OpenML R} bot benchmark data (final subset)}.
\newblock 2018.
\newblock URL
  \url{https://figshare.com/articles/OpenML_R_Bot_Benchmark_Data_final_subset_/5882230}.

\bibitem[Lang et~al.(2017)Lang, Bischl, and Surmann]{Lang2017}
M.~Lang, B.~Bischl, and D.~Surmann.
\newblock batchtools: Tools for {R} to work on batch systems.
\newblock \emph{The Journal of Open Source Software}, 2\penalty0 (10), 2017.

\bibitem[Meyer et~al.(2017)Meyer, Dimitriadou, Hornik, Weingessel, and h]{svm}
D.~Meyer, E.~Dimitriadou, K.~Hornik, A.~Weingessel, and F.~L. h.
\newblock \emph{e1071: Misc Functions of the Department of Statistics,
  Probability Theory Group (Formerly: E1071), TU Wien}, 2017.
\newblock URL \url{https://CRAN.R-project.org/package=e1071}.
\newblock R package version 1.6-8.

\bibitem[Probst et~al.(2018)Probst, Bischl, and Boulesteix]{Probst2018}
P.~Probst, B.~Bischl, and A.-L. Boulesteix.
\newblock Tunability: Importance of hyperparameters of machine learning
  algorithms.
\newblock \emph{ArXiv preprint arXiv:1802.09596}, 2018.
\newblock URL \url{https://arxiv.org/abs/1802.09596}.

\bibitem[Reif(2012)]{Reif2012ACD}
M.~Reif.
\newblock A comprehensive dataset for evaluating approaches of various
  meta-learning tasks.
\newblock In \emph{ICPRAM}, 2012.

\bibitem[Schliep and Hechenbichler(2016)]{kknn}
K.~Schliep and K.~Hechenbichler.
\newblock \emph{kknn: Weighted k-Nearest Neighbors}, 2016.
\newblock URL \url{https://CRAN.R-project.org/package=kknn}.
\newblock R package version 1.3.1.

\bibitem[Smith et~al.(2014)Smith, White, Giraud-Carrier, and
  Martinez]{Smith2014}
M.~R. Smith, A.~White, C.~Giraud-Carrier, and T.~Martinez.
\newblock An easy to use repository for comparing and improving machine
  learning algorithm usage.
\newblock In \emph{Meta-Learning and algorithm selection workshop at ECAI
  2014}, page~41, 2014.

\bibitem[Therneau and Atkinson(2018)]{rpart}
T.~Therneau and B.~Atkinson.
\newblock \emph{rpart: Recursive Partitioning and Regression Trees}, 2018.
\newblock URL \url{https://CRAN.R-project.org/package=rpart}.
\newblock R package version 4.1-12.

\bibitem[{van Rijn} and {Hutter}(2017)]{Rijn2017}
J.~N. {van Rijn} and F.~{Hutter}.
\newblock {Hyperparameter importance across datasets}.
\newblock \emph{ArXiv preprint arXiv:1710.04725}, 2017.
\newblock URL \url{https://arxiv.org/abs/1710.04725}.

\bibitem[Vanschoren et~al.(2013)Vanschoren, van Rijn, Bischl, and
  Torgo]{OpenML2013}
J.~Vanschoren, J.~N. van Rijn, B.~Bischl, and L.~Torgo.
\newblock {{OpenML}: Networked Science in Machine Learning}.
\newblock \emph{SIGKDD Explorations}, 15\penalty0 (2):\penalty0 49--60, 2013.

\bibitem[Wright and Ziegler(2017)]{ranger}
M.~N. Wright and A.~Ziegler.
\newblock {ranger}: A fast implementation of random forests for high
  dimensional data in {C++} and {R}.
\newblock \emph{Journal of Statistical Software}, 77\penalty0 (1):\penalty0
  1--17, 2017.

\end{thebibliography}

\end{document}